\documentclass[10pt]{article}
\usepackage{spconf,amsfonts,amsmath,mathtools,graphicx,color}

\usepackage{txfonts}                    
\usepackage[T1]{fontenc}
\usepackage{pstricks,pstricks-add,pst-all,psfrag}
\psset{unit=1mm,linewidth=.5pt,arrowscale=2,nodesep=0mm,labelsep=0mm}
\usepackage[sort, numbers]{natbib}

\usepackage{tikz}
\usetikzlibrary{arrows,chains,matrix,positioning,scopes}
\usepackage{subfigure}

\newcommand*{\ul}{\underline}

\newcommand*{\inp}{\ul{x}}

\newcommand*{\x}{\ul{x}_{l}}
\newcommand*{\xq}{\ul{x}_{l-1,q}}

\newcommand*{\lact}{\ul{a}_{l}}
\newcommand*{\lactq}{\ul{a}_{l,q}}
\newcommand*{\w}{{\bf W}_{l}}
\newcommand*{\wq}{{\bf W}_{l,q}}
\newcommand*{\bi}{\ul{b}_{l}}
\newcommand*{\bq}{\ul{b}_{l,q}}

\newcommand{\floor}[1]{ \left \lfloor #1 \right \rfloor}

\begin{document}

\title{Deep Neural Network inference with reduced word length}

\name{Lukas Mauch, Lukas Enderich and Bin Yang}
\address{Institute of Signal Processing and System Theory,
         University of Stuttgart, Germany}
\maketitle

\begin{abstract}
    Deep neural networks (DNN) are powerful 
    models for many pattern recognition tasks,
	yet their high computational complexity 
	and memory requirement limit them to applications on high-performance
	computing platforms. In this paper, we propose a new method to evaluate DNNs
	trained with $32 \: \mathrm{bit}$ floating point (float32) accuracy
	using only low precision integer arithmetics in combination with binary 
	shift and clipping operations. 
	Because hardware implementation of these operations is much simpler than 
	high precision
	floating point calculation, our method can be used for an efficient DNN inference
	on dedicated hardware. In experiments on MNIST, we demonstrate that DNNs trained with 
	float32 can be evaluated using a combination 
	of $2 \: \mathrm{bit}$ integer arithmetics and a few float32 calculations in each layer 
	or only $3 \: \mathrm{bit}$ integer arithmetics 
	in combination with binary shift and 
	clipping without significant
	performance degradation.
\end{abstract}


\section{Introduction}
    Today, deep neural networks (DNN) are applied to many machine learning tasks with great success. They 
    are suited to solve many problems from image recognition, segmentation and 
	natural language processing \cite{LiDeng14,dlReview}, delivering state-of-the-art results. 
	Their success is mainly based on improved training methods, 
	their ability to learn complex relationships directly from raw input 
	data and on the increasing computational power that makes 
	processing of large datasets and training of large models with up to millions of parameters possible. 
	Despite their success, DNNs are not suited for mobile applications 
	or the use on embedded devices because they are computationally complex and 
	need large memory to store the huge number of parameters. 
	Network reduction methods address this problem and reduce the computational and memory complexity of
	DNNs without significant performance degradation.
	
	Consider a generic dense feed-forward network with $L$ layers that computes
	\begin{eqnarray}
        \lact &=& \w \ul{x}_{l-1} + \bi  \label{eq:activation} \\
        \x &=& \Phi_l (\lact), \label{eq:output}
        \end{eqnarray}
	where $l=1,...,L$ is the layer index and $\ul{x}_{l-1} \in \mathbb{R}^{M_{l-1}}$, $\lact \in \mathbb{R}^{M_l}$ and $\x$ $\in \mathbb{R}^{M_l}$ 
	are the input, activation and output of a layer, respectively. 
	
	The memory complexity of layer $l$ is 
	$C^l_{mem} = (M_{l-1}M_l + M_l) K \: \mathrm{bits}$ to store the
	elements of the weight matrix $\w \in \mathbb{R}^{M_l \times M_{l-1}}$ and the bias vector $\bi \in \mathbb{R}^{M_l}$ using
	a word length of $K \: \mathrm{bits}$. 
	The parameters are typically stored in $32 \: \mathrm{bit}$ floating point (float32) format with $K=32$. 
	The computational complexity consists of $C^l_{mac} = M_lM_{l-1}$
	multiplication and accumulation (MAC) operations. 

    One approach to reduce the computational and memory 
    complexity of trained DNNs is to reduce 
    the number of parameters in the network. 
    Examples are factorization or pruning methods, which 
    either exploit the low-rank or the sparse structure of the weight matrices $\w$ 
    \cite{zhang_cnn14, zhang_cnn15, mauch_pruning_17, mauch_factors_17, braindamage}.     
    
    Another approach is to quantize all parameters of a DNN and to map each parameter of a DNN
    to the nearest element of a small set of values \cite{Han15}. 
    Prior works demonstrated that clustering techniques can be used to 
    find such a small set of values that approximate the parameters of a trained DNN with high accuracy.
    DNNs proved to be very robust to quantization, meaning that only few cluster centers are needed
    to represent the parameters of a DNN. Because only the cluster centers must be stored in float32 and
    because the cluster index of each parameter can be encoded
    with only few bits, quantization considerably reduces the memory complexity
    of trained DNNs. Others used fixed point quantization to reduce the memory complexity \cite{Lin15}. 
    However, during inference the network is still evaluated with floating point arithmetics.
    
    Vanhoucke et. al proposed a method to speed up DNN inference
    using quantization and a combination of fixed point and float32 arithmetics \cite{van11}.
    We propose a similar, but new method to evaluate DNNs which uses only
    low precision integer arithmetics, binary shift 
    and clipping operations and does not need any float32 calculations.
    
    We apply
    uniform quantization to convert the float32 inputs and parameters of
    a network to low precision integer values. The 
    activation of each network layer is evaluated using
    only low precision integer calculations. For networks with 
    arbitrary activation functions, the layer activation must be converted back 
    to float32 to evaluate the activation function, resulting in 
    mixed integer/float32 computations. For networks with relu activation functions, 
    we demonstrate that the network can be evaluated using only 
    integer arithmetics which are much simpler to implement in hardware. 
    Therefore, our method 
    is the key for an efficient implementation of DNNs on dedicated hardware.
    Special hardware can, for example, perform multiple low precision integer operations in the same time
    as one floating point operation. 
    For example, NVIDIAs recent Pascal architecture can perform
    4 int8 MACs in the same time as a single float32 multiplication. This means, a DNN
    can be evaluated up to 4 times faster.
    
    The contributions of this paper are: 
    1) We propose a new method to evaluate DNNs that uses uniform 
    quantization and only needs low precision integer arithmetics, binary shift and
    clipping operations.
    2) We discuss how training with regularization influences the distribution of
    the parameters of a DNN and thus the performance of the quantized DNN.

 \section{Uniform quantization}   \label{sec:quantization}
\begin{figure}
            \centering
            \includegraphics[scale=.37]{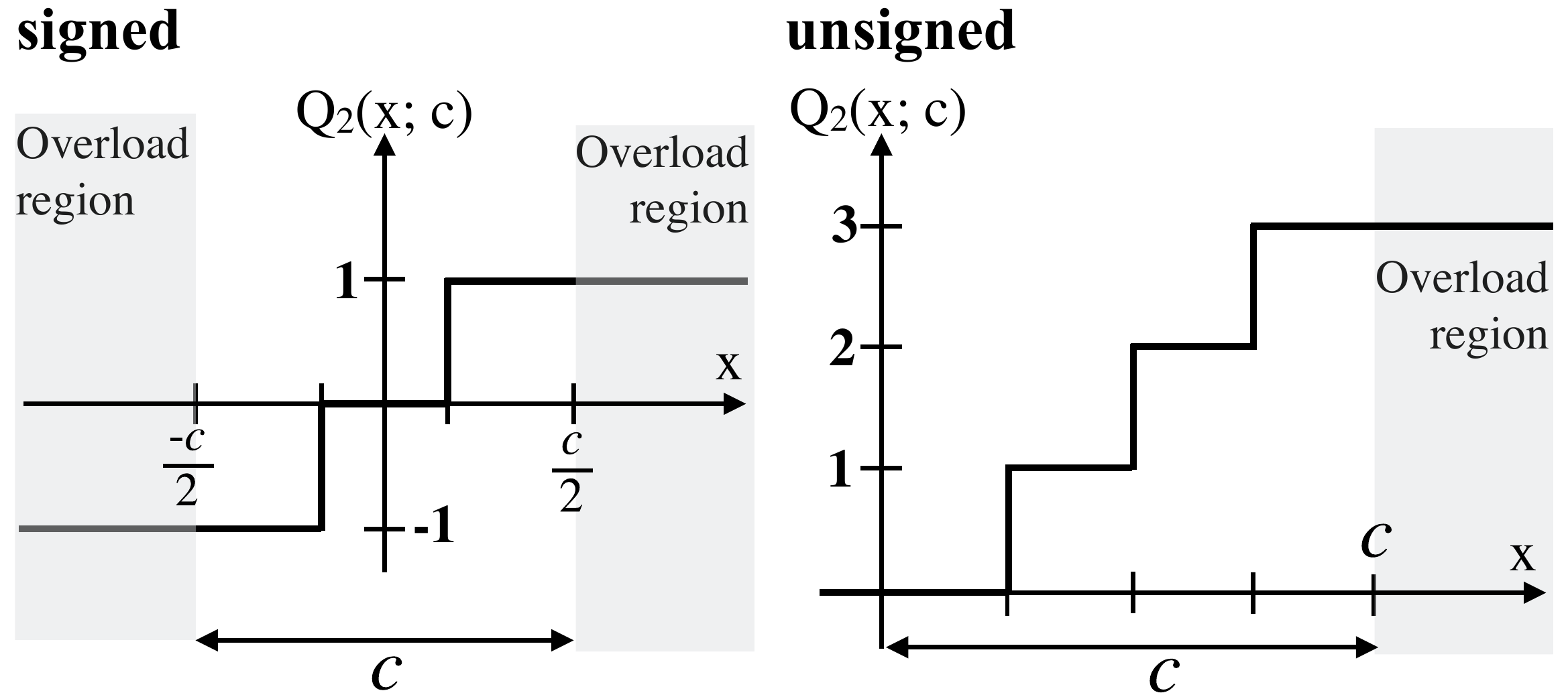}
            \vspace{-20pt}
            \caption{Uniform quantization for unsigned and signed values with $K=2$.}
            \label{fig:quantization}
        \end{figure}

        Consider a uniform quantization function $Q_K(x;\Delta)$ 
        that maps a value $x$ to an integer value of word length $K$.
        As shown in Fig. \ref{fig:quantization}, we distinguish between quantization of signed and unsigned values.
        
        For signed values $x \in \mathbb{R}$, we use a quantization function $Q_K: \mathbb{R} \rightarrow \mathbb{Z}_K$ 
        to map real values from the original value range $[-c/2, c/2]$ to a set of $2^K-1$ 
        signed integer values that are symmetrically distributed around $0$, i.e. $\mathbb{Z}_K = \{ -2^{K-1}+1, ..., 2^{K-1}-1 \}$.
        We define the signed quantization function $Q_K(x;\Delta)$ as
        \begin{equation}
            x_q = Q_K(x;\Delta) = \mathrm{clip}_{K} \left ( \floor{\frac{x}{\Delta}+\frac{1}{2}} \right )
        \end{equation}
        $\Delta=\frac{c}{2^K-1}$ is the step width
        \begin{equation}
            \mathrm{clip}_K(x) = \begin{cases}
                                    x &, |x| \leq 2^{K-1}-1\\
                                    -2^{K-1}-1 &, x<-2^{K-1}+1\\
                                    2^{K-1}-1 &, x>2^{K-1}-1
                                \end{cases}
        \end{equation}
        is the clipping (saturation) of the quantization function, which maps all values in the overload region to
        the maximum or minimum value in $\mathbb{Z}_K$.
        For unsigned values $x \in \mathbb{R}^+$, we use a quantization function $Q_K: \mathbb{R}^+ \rightarrow \mathbb{Z}^+_K$
        to map real values from the value range $[0,c]$ to a set of $2^K$ 
        unsigned integer values $\mathbb{Z}^+_K = \{ 0, ..., 2^{K}-1 \}$.
        We define the unsigned quantization function $Q_K(x;\Delta)$ as
        \begin{equation}
            x_q = Q_K(x;\Delta) = \floor{\frac{x}{\Delta}}
        \end{equation}
        The step width is $\Delta=\frac{c}{2^K}$ and the clipping function is
        \begin{equation}
            \mathrm{clip}_K(x) = \begin{cases}   
                                    x &, x \leq 2^K-1\\
                                    2^{K}-1 &, x>2^K-1
                                    \end{cases}.
        \end{equation}
        
        Quantization introduces the quantization noise $\epsilon$, meaning $x = \Delta x_q + \epsilon$.
        For given $K$, the step size $\Delta$ must be chosen to balance high resolution (small $\Delta$) against
        small overload regions (large $\Delta$).
        One approach is to minimize the mean square quantization error
        $E[\epsilon^2]$,
        which depends on the probability density function $p(x)$ of $x$. 
        A compact $p(x)$ with short
        tails is desirable to minimize $E[\epsilon^2]$.       
        
        Using uniform quantization, a function $y=\ul{w}^T\ul{x} + b$ 
        with $\ul{w}, \ul{x} \in \mathbb{R}^N$ and $b \in \mathbb{R}$ is approximately
        \begin{equation}
            y \approx \Delta_w \Delta_x \left (\sum_{i=1}^N  Q_K(w_i;\Delta_w)Q_K(x_i;\Delta_x) + Q_K(b;\Delta_w\Delta_x) \right ),
        \end{equation}
        where all terms within the bracket can be computed with integer arithmetics
        of word length $K$. If $\Delta_w \Delta_x$ is also chosen as the step size for quantization of $b$, 
        only one single floating point multiplication $\Delta_x \Delta_w$ is needed
        to map the result $y_q$ back to the original value range.
        
        We assume that integer multiplications produce no overflows. This means, 
        a multiplication of integer values of word legnth $K$ results in an integer
        of twice the word length ($\mathbb{Z}_K \times \mathbb{Z}_K \rightarrow \mathbb{Z}_{2K}$ 
        and $\mathbb{Z}_K^+ \times \mathbb{Z}_K \rightarrow \mathbb{Z}_{2K}$).

 \section{Quantization of layers}
        Of course, the word length does not increase from layer to layer. 
        The input of each layer is quantized from $\mathbb{Z}_{2K}$ back to $\mathbb{Z}_{K}$.
        After training, the layers of a DNN can be converted to quantized 
        layers which only use low precision integer arithmetics
        to calculate the layer activation. 
        Although only dense layers are considered below, the
        idea can be easily adapted to convolutional layers as well.
        
\begin{figure}
            \centering
            \includegraphics[scale=.3]{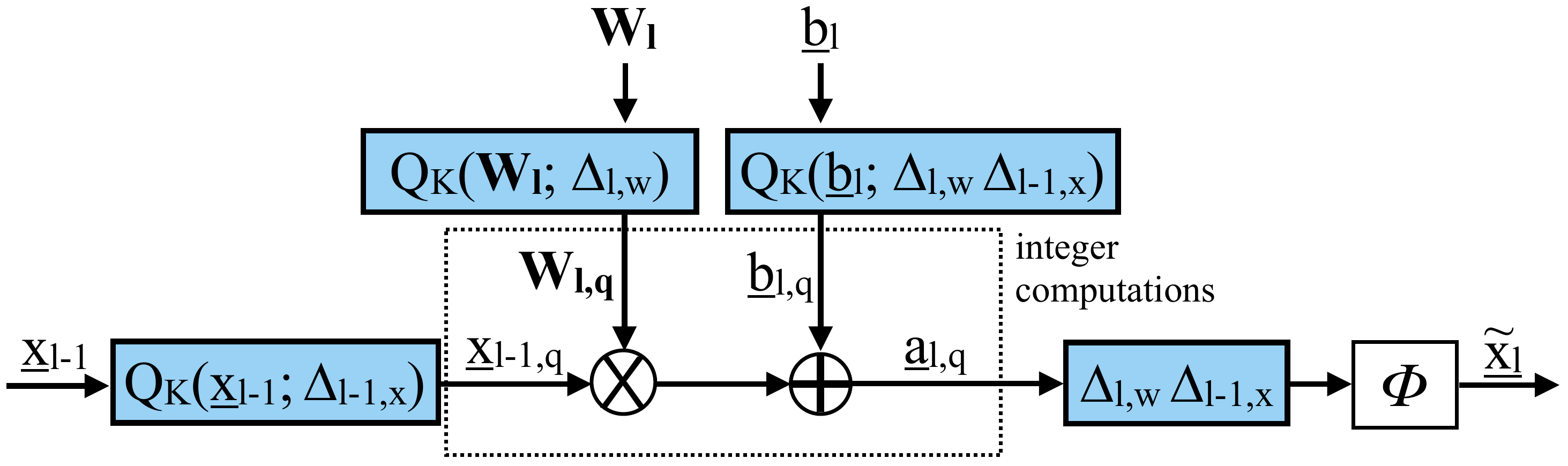}
            \vspace{-25pt}
            \caption{Using low precision integer arithmetics to evaluate a layer}
            \label{fig:quantization_layer}
        \end{figure}

        As shown in Fig. \ref{fig:quantization_layer}, 
        the input $\ul{x}_{l-1}$ and the 
        parameters $\w$ and $\bi$ 
        are quantized to vectors and matrices containing 
        integer values of word length $K$. 
        $\xq=Q_K(\ul{x}_{l-1};\Delta_{l-1,x})$, $\wq=Q_K(\w;\Delta_{l,w})$ and $\bq=Q_K(\bi ;\Delta_{l-1,x}\Delta_{l,w})$ are quantized
        with uniform quantization functions which 
        are defined elementwise. We use signed quantization for the parameters
        $\w$ and $\bi$ and unsigned quantization for $\x$, 
        since we assume unsigned $\x$ in each layer. This is feasible for networks with relu or sigmoid activation functions.

        The quantized layer computes 
        \begin{eqnarray}
            \lactq &=& \wq \xq + \bq \\
            \tilde{\ul{x}}_l &=& \Phi_l ( \Delta_{l-1,x} \Delta_{l,w} \lactq ) = \x + \ul{\epsilon}_{l,K},
        \end{eqnarray}
        where $\ul{\epsilon}_{l,K} \in \mathbb{R}^{M_l}$ is the quantization noise. In general, a different word length $K$ can be chosen for 
        $\ul{x}_{l-1}$, $\w$ and $\bi$, in each layer. However, in each layer, we use the same fixed $K$ for the inputs and parameters and only choose $\Delta_{l-1,x}$,
        $\Delta_{l,w}$ and $\Delta_{l,b}$ such that the quantization error $\ul{\epsilon}_{l,K}$ is minimized in each layer, seperately.
        For this purpose, we use $N$ training samples $\{\inp(n)\}_{n=1}^N$ to estimate the mean square quantization error $E[||\ul{\epsilon}_{l,K}||^2]$ and minimize it
        using a grid search. 
        
        We use $\Delta_{l,b} = \Delta_{l-1,x} \Delta_{l,w}$ as the step size to quantize $\bi$. As discribed in section \ref{sec:quantization}, 
        the activation of the layer can be converted back to the original value range
        using just $M_l$ floating point multiplications. This overhead is small compared to the $M_lM_{l-1}$ multiplications, which can now
        be evaluated with low precision integer arithmetics.
    
        \begin{figure}
            \centering
            \includegraphics[scale=.3]{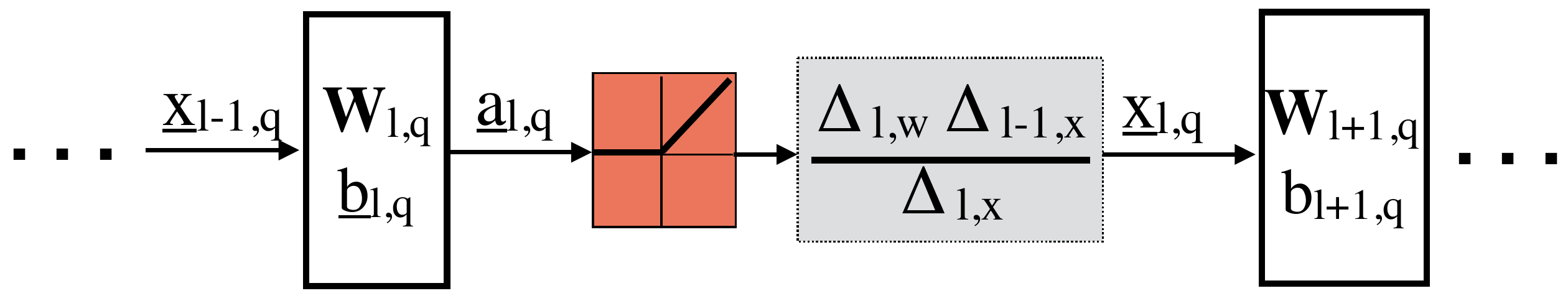}
            \vspace{-15pt}
            \caption{Using low precision integer arithmetics and binary shift to evaluate a network with relu activation function}
            \label{fig:quantization_relu}
        \end{figure}
        
        If each layer of the DNN uses a relu activation function
        and if we choose $\Delta_{l-1,x}$ and $\Delta_{l,w}$ as powers of 2 
        (i.e. $\Delta_{l-1,x},\Delta_{l,w} \in \{2, 4, 8,...\}$), the computation 
        of the network output can be further simplified, as shown in Fig. \ref{fig:quantization_relu}.
        Since $\Phi_l(\cdot)$ is piecewise linear and $\Phi_l(\Delta_{l-1,x}\Delta_{l,w}\lactq) = \Delta_{l-1,x}\Delta_{l,w}\Phi_l(\lactq) \in \mathbb{Z}_{3K}^M$,
        we can compute the quantized input of layer $l+1$ by
        \begin{equation}
            \ul{x}_{l,q} = \mathrm{clip}_K \left ( \frac{\Delta_{l-1,x} \Delta_{l,w}}{\Delta_{l-1,x}} \Phi_l(\lactq) \right ). 
        \end{equation}
        This means, we only need to apply a binary shift
        followed by a clipping to each element of $\Phi_l(\lactq)$.
        Because no floating point
        multiplications are needed, the DNN can be 
        evaluated using only low precision integer arithmetics.

\section{Regularization and quantization}
        Assume that we want to quantize $x \in \mathbb{R}$ with $x \sim p(x)$. 
        As described in secion \ref{sec:quantization},
        this introduces a quantization error $\epsilon$, which depends on the shape of $p(x)$. If $x$ has a high probability to have values within the overload region $|x|>1/2$, 
        the mean squared quantization error $E[\epsilon^2]$ will be large. Therefore, a compact distribution $p(x)$ with no tails 
        ($p(x) = 0, \:\: \forall |x|> c/2$) is desirable for a small quantization error.
        
        During training, regularization can control the shape of the distribution of the parameters $\w$ and $\bi$. 
        A common way to regularize a network is to augment the cost function with a regularization term $\lambda ||\w||_p^p, \:\: p = 1,2,...$,
        where $\gamma$ is the regularization parameter.
        For $p=1$, the regularization term acts like a Laplacian prior for $\w$, promoting sparse weight matrices with long tails in distribution. 
        This is the desired way to regularize a DNN if 
        pruning is used for network reduction. After training, small elements in $\w$
        can be set to zero to remove (prune) connections from the network. However, it is not desirable for quantization.
        A regularization term with $p\rightarrow \infty$ penalizes the largest absolute values in $\w$ and thus promotes weight matrices
        with a compact distribution $p([\w]_{ij})$. This is desired for quantization. 
        Our experiments show that a careful choice of $p$ during training is the key to a good 
        model accuracy after quantization.

\section{Experiments}    
        All experiments are done using Theano \cite{theano} and Keras \cite{keras}.
        We use network quantization with two networks trained on MNIST dataset \cite{mnist}
        which contains gray-scale images of pre-segmented handwritten digits that are 
        divided into a training set with 60000 and test set with 10000 images of size 28x28.
        
        The MNISTnet1 is a dense DNN with $L=3$ layers and $512/512/10$ neurons. 
        The MNISTnet2 is a convolutional neural network with $L=6$ layers. 
        We use $4$ convolutional layers with kernel size $3 \times 3$ and $16/16/32/32$ feature maps.
        Maxpooling with stride $2\times2$ is applied after the layers $2$ and $4$. 
        The two dense output layers of MNISTnet2 contain $128$ and $10$ neurons.
        
        For both networks, we use relu activation functions in all hidden layers
        and a softmax activation function with cross-entropy loss in the output layer. 
        The networks are optimized with Adam for $30$ epochs, using
        dropout with probability $p=0.1$. 
        After training, MNISTnet1 and MNISTnet2 achieve the baseline accuracies $98.4\%$ and $99.15\%$, respectively.
        
        In our experiments, we apply network quantization with different
        word lengths $K=2,3,4,6,8$ and compare three different 
        quantization methods. These quantization methods use different approaches to
        choose the step sizes $\Delta_{l,w}$, $\Delta_{l,b}$ and $\Delta_{l-1,x}$.
        
        The first method uses the maximum
        absolute values from $\w$ and $\ul{x}_{l-1}$ to determine the step sizes 
        $\Delta_{l,w} = \mathrm{max}(|[\w]_{ij}|) \allowbreak / (2^K-1)$, $\Delta_{l,b} = \mathrm{max}(|[\bi]_{i}|) / (2^K-1)$ and
        $\Delta_{l-1,x} = \mathrm{max}(|[\ul{x}_{l-1}]_{i}|) \allowbreak / (2^K)$ for quantization. We call this method
        "max. absolute value". The second method
        chooses $\Delta_{l,w}$, $\Delta_{l,b}$ and $\Delta_{l-1,x}$ to minimize the mean squared quantization error in each layer, separately. 
        We call this method "min. MSE I".
        The third method also minimizes the quantization error, but restricts the step sizes
        $\Delta_{l,w},\Delta_{l,b},\Delta_{l-1,x} \in \{2,4,...\}$ to be powers of 2. We call this method "min. MSE II".
        This method is the most interesting one, because it allows to evaluate the network using only
        integer arithmetics, binary shift and clipping.
        
\begin{figure}
	  \centering
	   \subfigure[MNISTnet1 trained without regularization]{\includegraphics[scale=.65]{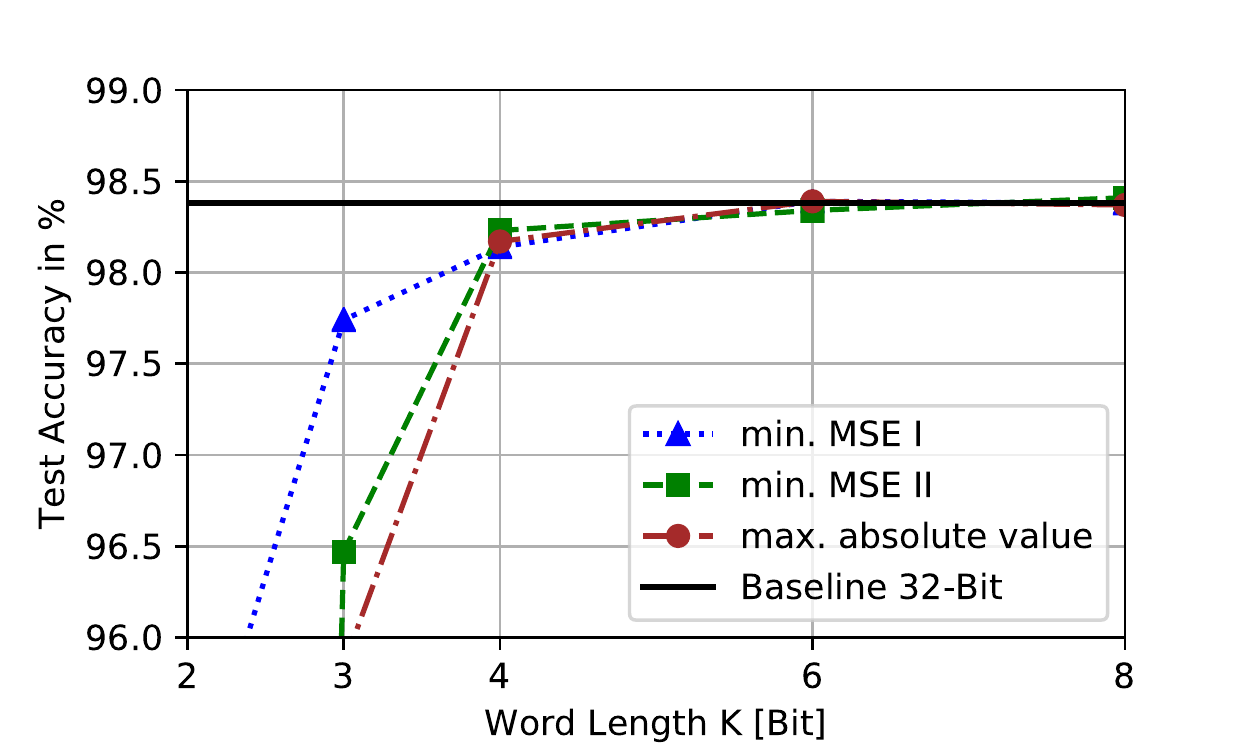}}\\[-1.5ex]
	   \subfigure[MNISTnet2 trained without regularization]{\includegraphics[scale=.65]{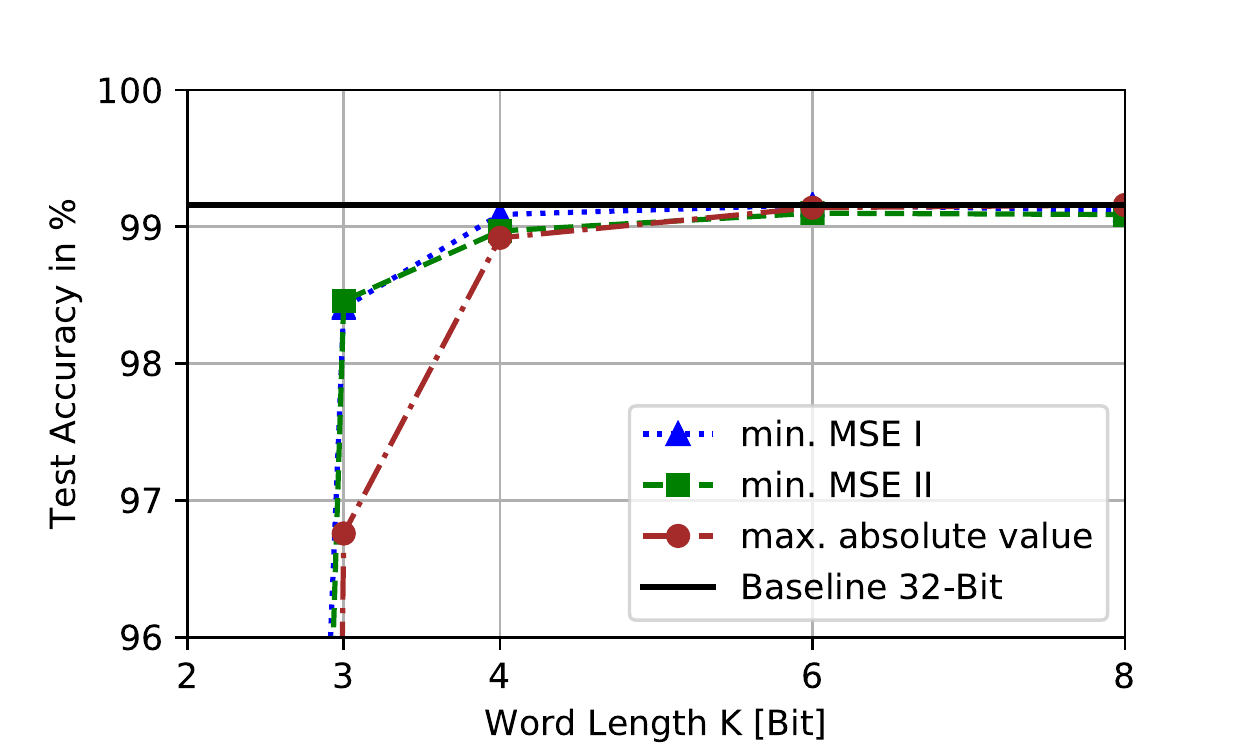}}\\[-1.5ex]
	   \subfigure[MNISTnet2, regularized with $p=1$]{\includegraphics[scale=.65]{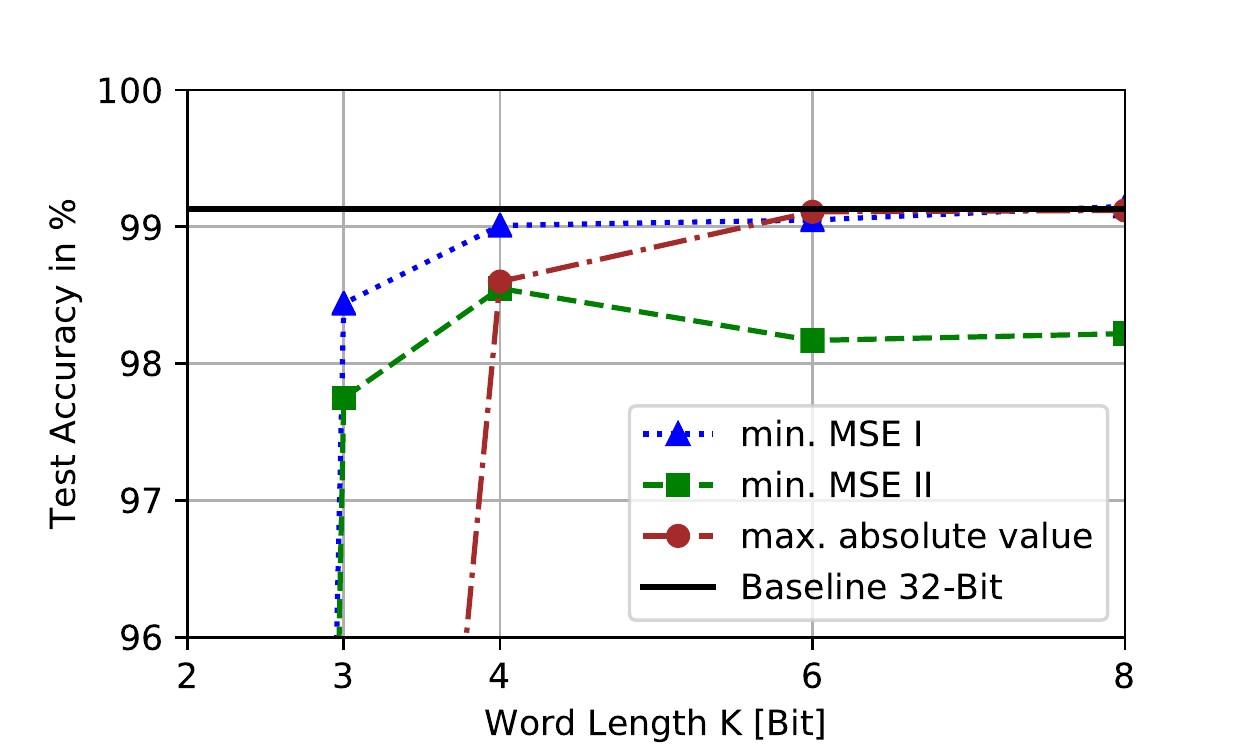}}\\[-1.5ex]
	   \subfigure[MNISTnet2, regularized with $p=8$]{\includegraphics[scale=.65]{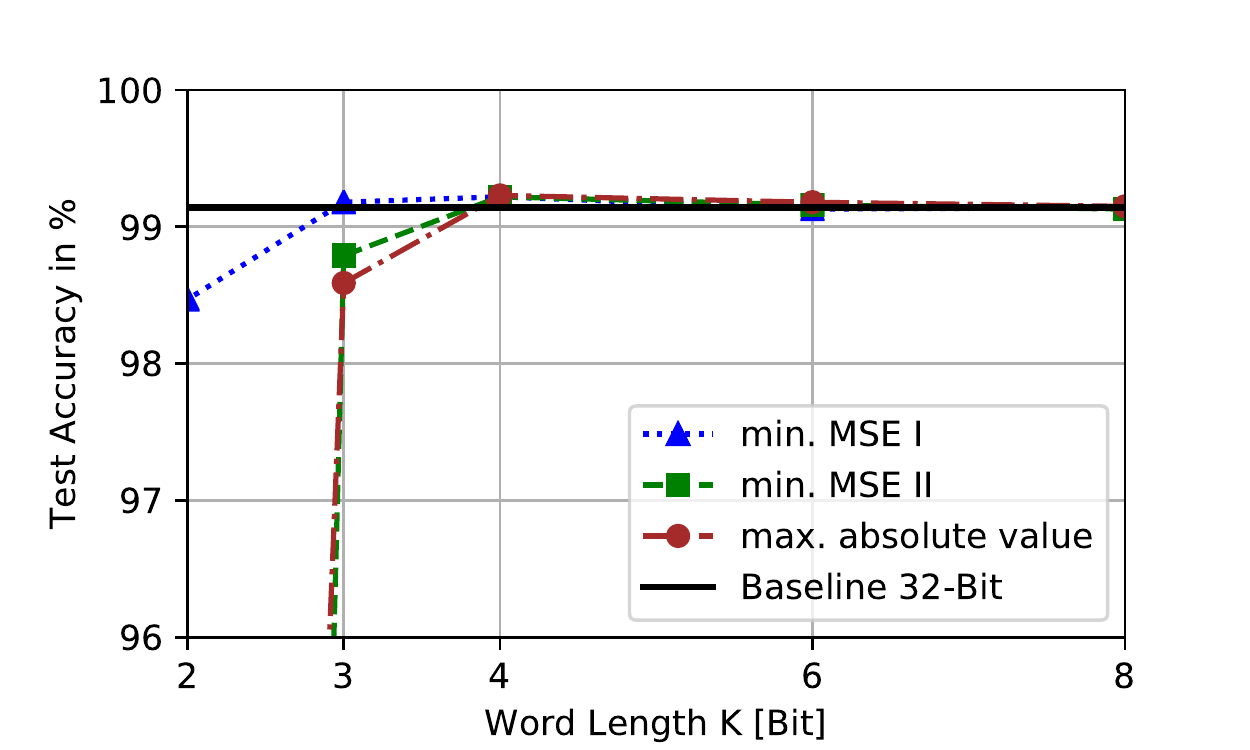}} \vspace{-10pt}
	  \caption{Quantization of MNISTnet1 and MNISTnet2 with $K=2,3,4,6,8$, after
	  training with different regularizers.}
	  \label{fig:result}
	\end{figure}

        For MNISTnet1, the results are shown in Fig. \ref{fig:result} (a).
        MNISTnet1 can be evaluated with $6 \: \mathrm{bit}$ integer arithmetics without considerable performance loss.
        For $K=2,3,4$, we observe that the quantized network performs 
        best if the step size for quantization is chosen to minimize the mean squared quantization error.
        
\begin{figure}
            \centering
            \includegraphics[scale=.3]{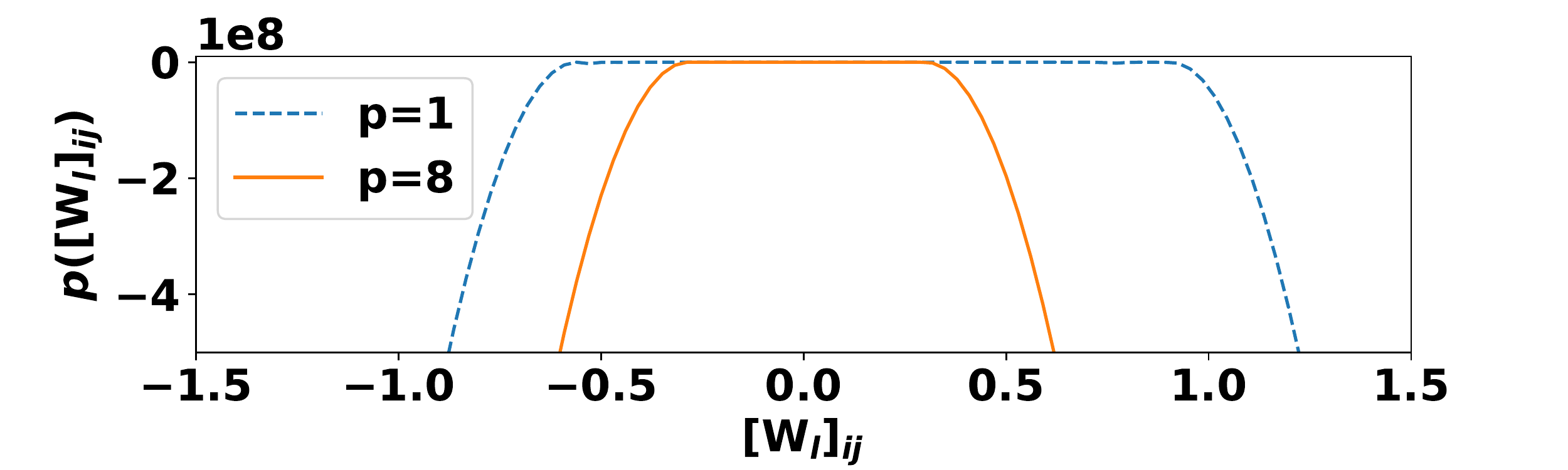}
            \vspace{-15pt}
            \caption{Distribution of the network weights after training for $p=1$ and $p=8$.}
            \label{fig:log_dense_weights}
        \end{figure}

        In our second experiment, we apply quantization to the MNISTnet2 using three different
        regularization methods during training, i.e. no regularization with $\lambda = 0$, 
        regularization with $\lambda = 1e-4$ and $p=1$ and with $\lambda = 1e-4$ and
        $p=8$. The results are shown in Fig. \ref{fig:result} (b) to \ref{fig:result} (d).
        We observe large accuracy degradations for $K=2,3,4$ if the MNISTnet2 is trained
        with $\lambda=1e-4$ and $p=1$. 
        For all three quantization methods, the accuracy of the quantized networks is even worse than 
        with MNISTnet2 trained without regularization ($\lambda=0$).
        However, this is expected since regularization with $p=1$ leads to a distribution
        $p([\w]_{ij})$ with long tails and therefore to large quantization errors. This is shown
        in Fig. \ref{fig:log_dense_weights}.
        
        Quantization works best if the MNISTnet2 
        is trained with $\lambda=1e-4$ and $p=8$. 
        Even with $2 \: \mathrm{bit}$ integer arithmetics the 
        accuracy of the MNISTnet2 is only reduced slightly if we use the quantization method "min. MSE I".
        This is astonishing since $K=2$ implies that only three weight values
        remain after quantization. For $K=3$, the quantization method "min. MSE II" 
        also leads to an accuracy very close to the baseline. This means, MNISTnet2 can be evaluated
        using only $3 \: \mathrm{bit}$ integer arithmetics, 
        binary shift and clipping operations.
       
        In general, quantization can be used in combination with other model
        reduction methods like factorization or pruning \cite{zhang_cnn14, zhang_cnn15, mauch_pruning_17}.
        However, we belief that one cause for the remarkable robustness of 
        DNNs to quantization noise is the large parameter redundancy in trained DNNs. If this redundancy
        is removed by other reduction methods, we belief that quantization will lead to larger performance degradation.
        The optimum trade-off between quantization and other model reduction methods is still an open issue.
        
\section{Conclusion}
	We proposed a method to quantize DNNs that was trained with floating point accuracy and
	to evaluate them using only low precision integer arithmetics. 
	In our experiments, we showed that our method leads to 
	almost no accuracy degradation and can therefore be used to implement trained 
	DNNs efficiently on dedicated hardware. 
	We also demonstrated how regularization during training influences the distribution of weights and thus network quantization.
	Our experiments show that $L_1$ regularization what is often used in combination with network pruning
	leads to DNNs that can not be quantized easily. Instead, regularizers
	that lead to a compact distribution of weights are beneficial for quantization.

\newpage
\bibliographystyle{IEEEbib}

\bibliography{article}

\end{document}